# Performance evaluation of Reddit Comments using Machine Learning and Natural Language Processing methods in Sentiment Analysis

Xiaoxia Zhang[1[0009-0007-5698-4677]], Xiuyuan Qi[2[0009-0005-9536-0870]] and Zixin Teng[3[0009-0009-2699-6743]]

Shanghaitech University, Shanghai, China
[1] zhangxx5@shanghaitech.edu.cn
[2] qixy1@shanghaitech.edu.cn
[3] tengzx@shanghaitech.edu.cn

**Abstract.** Sentiment analysis, an increasingly vital field in both academia and industry, plays a pivotal role in machine learning applications, particularly on social media platforms like Reddit. However, the efficacy of sentiment analysis models is hindered by the lack of expansive and fine-grained emotion datasets. To address this gap, our study leverages the GoEmotions dataset, comprising a diverse range of emotions, to evaluate sentiment analysis methods across a substantial corpus of 58,000 comments. Distinguished from prior studies by the Google team, which limited their analysis to only two models, our research expands the scope by evaluating a diverse array of models. We investigate the performance of traditional classifiers such as Naive Bayes and Support Vector Machines (SVM), as well as state-of-the-art transformer-based models including BERT, RoBERTa, and GPT. Furthermore, our evaluation criteria extend beyond accuracy to encompass nuanced assessments, including hierarchical classification based on varying levels of granularity in emotion categorization. Additionally, considerations such as computational efficiency are incorporated to provide a comprehensive evaluation framework. Our findings reveal that the RoBERTa model consistently outperforms the baseline models, demonstrating superior accuracy in fine-grained sentiment classification tasks. This underscores the substantial potential and significance of the RoBERTa model in advancing sentiment analysis capabilities.

## 1 Introduction

Sentiment analysis represents a pivotal domain within research, particularly within commercial contexts, owing to its significance in delineating user profiles on social platforms. Despite the considerable contributions of Natural Language Processing (NLP) researchers in developing sentiment classification datasets spanning various domains such as news [Strapparava and Mihalcea, 2007] and Twitter [Saif et al., 2013], these datasets frequently present limitations in both scale and granularity. Primarily, they adhere to binary or Ekman's emotion classification schemes[Ekman, 1992]. The binary taxonomy oversimplifies emotional nuances by categorizing sentiments solely into positive and negative types, thus lacking precision and accuracy. In the framework of discrete emotion theory, it is posited that all humans possess an intrinsic set of basic emotions that are universally recognizable across cultures, even though discussions regarding the classification of so-called "basic emotions" have been subject to ongoing controversy [Ortony, 2022]. The seminal work of Paul Ekman and his colleagues in their 1992 cross-cultural study delineated six fundamental emotions: anger, disgust, fear, happiness, sadness, and surprise, forming the classical Ekman's classification standard. However, the delineation of emotions into six categories fails to adequately address the intricate nature of contemporary emotion classification tasks, particularly within the dynamic landscape of social media where a vast array of emotions is expressed, often defying easy categorization. Consequently, there arises an urgent necessity for the development of a comprehensive, large-scale dataset capable of nuanced emotion classification, serving as the foundational premise of our study.

In addressing these challenges, we conducted a comprehensive evaluation of multiple datasets, ultimately selecting the GoEmotions dataset developed by Google as the foundational corpus for our study. Comprising 58,000 Reddit platform comments meticulously annotated to encompass 27 emotion categories or "neutral" sentiments, GoEmotions caters to the diverse array of information typified by Reddit as a comprehensive, large-scale social news forum. Diverging from Ekman's emotion classification taxonomy, GoEmotions incorporates a plethora of positive, negative, and ambiguous emotion categories, rendering it particularly suitable for downstream conversational understanding tasks necessitating nuanced emotion comprehension. Besides, the Reddit platform has hitherto not garnered significant attention from researchers in the field of sentiment recognition and prediction. The GoEmotions dataset, however, stands as the inaugural dataset specifically tailored to this platform, thereby stimulating our group members' interest in employing a wider array of machine learning models to train and evaluate classification results. They also train a bidirectional LSTM as an additional baseline but performs significantly worse than BERT.

[Demszky et al., 2020] presented detailed work for building the GoEmotions dataset and showing how to demonstrate the high quality of the annotations via Principal Preserved Component Analysis. They conduct transfer learning experiments with existing emotion benchmarks to prove that GoEmotions generalizes well to other domains and different emotion taxonomies. Furthermore, Google team tried to use a fine-tuned BERT based model originating from [Devlin et al., 2019] to test the dataset and achieved an average F1-score of .46 for the proposed taxonomy of 27 emotion categories, .64 for an Ekman grouped model and .69 for a sentiment-grouped model. Meanwhile, the bidirectional LSTM only obtained an average F1-score of .41 for the full taxonomy, .53 for an Ekmangrouped model and .6 for a sentiment-grouped model.

Based on the research conducted by the Google team and the comprehensive preprocessing efforts applied to this dataset, we intend to utilize the results obtained from this dataset in conjunction with the Google BERT model as the baseline for our model evaluation. Additionally, we aim to initially implement and evaluate the performance of various Bayesian models and Support Vector Machines (SVMs), among other machine learning models covered in the curriculum. Furthermore, we will focus on large language models based on the RoBERTa model and the GPT interface. Our objective is to iteratively fine-tune and assess the classification accuracy, computational efficiency, and other relevant metrics of these models, with the ultimate aim of achieving superior training outcomes compared to the original Google models.

# 2 Methodology

## 2.1 Machine Learning Method

Our evaluation process commences with the utilization of the GoEmotions dataset. We aim to train a variety of machine learning models covered in our curriculum, alongside advanced natural language processing models. These models encompass a comprehensive pipeline comprising tokenization, embedding, transformation, and post-processing steps, such as applying sigmoid activation functions. Subsequently, all models undergo one-hot encoding to generate a series of outcomes, which are evaluated using diverse performance metrics. The specific workflow of this process is depicted in the diagram below.

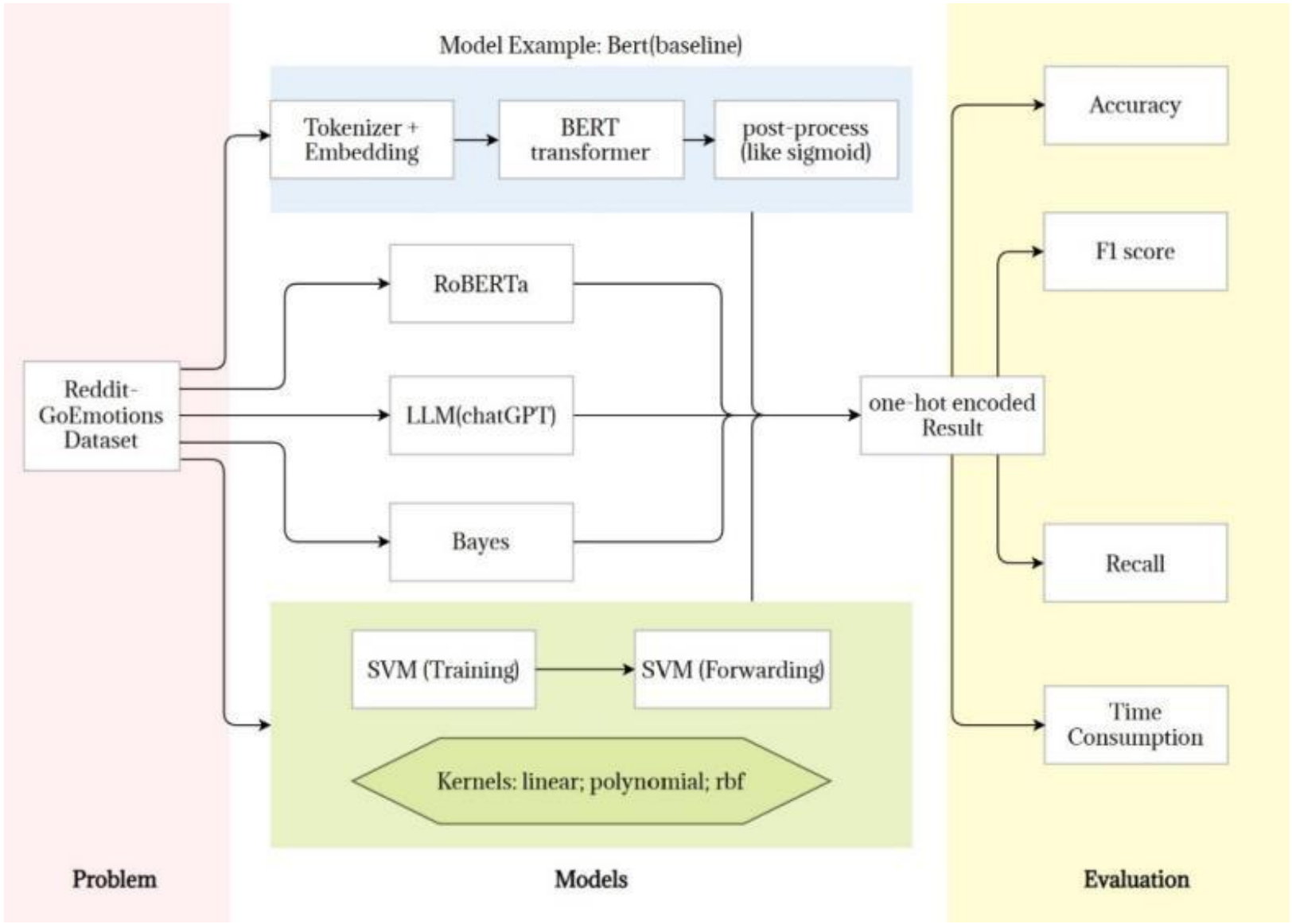

**Fig. 1.** Workflow of the process of our evaluation study

**Bayes + MLE/KNN** The Bayesian model, based on Bayesian probability theory, provides a probabilistic framework for sentiment classification tasks. It calculates sentiment label probabilities using Bayesian inference, enabling principled decision-making. By addressing uncertainty, it offers interpretability and resilience, valuable for text data analysis. Integrating Bayesian modeling with Maximum Likelihood Estimation (MLE) and K-Nearest Neighbors (KNN) methods enhances our approach, particularly in the Complement Naive Bayes Classifier.

**SVM** In sentiment classification prediction, Support Vector Machine (SVM) is a prominent supervised learning method widely used in text analysis tasks. SVM maps data points into high-dimensional spaces to find an optimal hyperplane for classification. Its role involves effectively discerning sentiment orientations in text, handling high-dimensional feature spaces, and exhibiting resilience against noise. By utilizing different kernel functions like linear, polynomial, and radial basis function (RBF), SVM adapts to diverse feature mappings for tailored sentiment analysis, enhancing classification accuracy. We will test all the above three kernels.

### 2.2 Natural Language Processing Method

**BERT** The BERT (Bidirectional Encoder Representations from Transformers) model serves as our baseline, representing a focal point of research by the Google team. They enhanced the model by incorporating a dense output layer to facilitate multi-label classification, accompanied by the utilization of a sigmoid cross-entropy loss function. Google emphasized the importance of training for a minimum of 4 epochs to effectively capture the underlying data patterns, while cautioning against the risk of overfitting with prolonged training durations. Leveraging the fine-tuned hyperparameters established by [Devlin et al., 2019] alongside the adjustments recommended by Google for the batch size set at 16 and a learning rate of 5e-5, is anticipated to yield optimal performance in the GoEmotions dataset.

**RoBERTa** The RoBERTa model, proposed in [Liu et al., 2019], is an extension of Google's BERT model introduced in 2018. Building upon BERT, RoBERTa modifies crucial hyperparameters, such as removing the next-sentence pretraining objective and employing larger mini-batches and learning rates during training. This refinement aims to enhance the model's robustness and performance. In sentiment classification tasks, RoBERTa's enhanced pretraining strategies and robust architecture may offer improved capabilities for accurately capturing nuanced sentiment expressions and achieving state-of-the-art performance in sentiment classification benchmarks.

**LLM(GPT)** The GPT (Generative Pre-trained Transformer) model, including GPT-3.5 Turbo from OpenAI, is a cutting-edge language generation model based on transformer architecture. Unlike BERT, which focuses on bidirectional contextual understanding, GPT generates text sequentially, predicting the next token based on preceding ones. Although not originally designed for sentiment classification, GPT can be adapted for such tasks by conditioning on sentiment-related prompts. In sentiment analysis, GPT generates text expressing sentiment, which can then be categorized. We'll implement a framework using GPT-3.5 for text sentiment analysis and evaluate its performance.

## 3 Experiment

### 3.1 Dataset

As we mentioned in introduction, the GoEmotions dataset is the largest manually annotated dataset of 58k English Reddit comments, labeled for 27 emotion categories or Neutral. The training dataset employed for model training comprises instances where a consensus exists among at least 2 raters. These datasets have no header row and are structured with three columns: "text," representing the textual content; a comma-separated list denoting emotion IDs, indexed from 0 to 27; and an identifier for each comment. Please note that for the same text, it may belong to more than one emotional category; therefore, it can have multiple emotion IDs associated with it.

Specifically, the emotion categories are: admiration, amusement, anger, annoyance, approval, caring, confusion, curiosity, desire, disappointment, disapproval, disgust, embarrassment, excitement, fear, gratitude, grief, joy, love, nervousness, optimism, pride, realization, relief, remorse, sadness, surprise. They are indexed by integers 0 to 26. And the number 27 stands for the nuetral type.

**Table 1.** Data example for original task.

| Text | Emotion IDs | Comment ID |
|---|---|---|
| I'm glad you had a great time here! Wishing you safe travels! | 17,20 | ee8f04x |
| Yes, he did. | 4,27 | ee1vzq2 |

Let's take a look at the examples in Table 1. For the first data entry, the emotion IDs are 17 and 20, indicating that the emotion classification labels for this sentence are "joy" and "optimism". For the second data entry, the emotion IDs are 4 and 27, indicating that the emotion classification labels for this sentence are "approval" and "neutral". It is evident that non-unique emotion IDs are more congruent with the authentic linguistic contexts encountered in daily life. This, in turn, imposes heightened classification requirements on the model's capabilities.

The GoEmotions dataset offers several advantages. Firstly, a significant majority of the examples (83%) possess a singular emotion label, ensuring clarity in annotation. Additionally, a high level of agreement among raters (94%) on these single labels enhances the dataset's reliability. To further enhance data quality, emotion labels selected by only one annotator are filtered out, resulting in the retention of 93% of the original data. Finally, the dataset is thoughtfully partitioned into train (80%, 43410 samples), dev (10%, 5426 samples), and test (10%, 5427 samples) sets, facilitating robust model evaluation. In addition to the 28 finer-grained sentiment categories (referred to as 'the original task'), we've kept Ekman's six-category classification and grouped sentiment into positive/negative/neutral/ambiguous categories. This enables evaluating models' abilities across different levels of classification precision. For these two types of classification tasks, the dataset maintains the same structure, with only the index of the emotion ID being altered. For the Ekman emotion classification task, IDs 0 through 6 respectively represent: anger, disgust, fear, joy, neutral, sadness, surprise; whereas for the general group clustering task, IDs 0 through 3 respectively represent: ambiguous, negative, neutral, positive.

### 3.2 Result

**Evaluation Metrics** We will use four metrics to test our models: Accuracy, F1 Score, Recall and Time Consumption. And we will comprehensively consider both weighted and macro metrics to ensure the completeness of the evaluation results. The accuracy, in the context of classification tasks, refers to the proportion of correctly classified instances out of the total number of instances evaluated.

$$Accuracy = \frac{\text{Number of Samples with Correct Classification}}{\text{Total Number of Samples}} \#(1)$$

Precision measures the proportion of true positive predictions among all positive predictions made by the model. And recall measures the proportion of true positive predictions among all actual positive instances in the dataset. The definition of F1 Score is:

$$\text{F1 Score} = 2 \times \frac{\text{Precision} \times \text{Recall}}{\text{Precision} + \text{Recall}} \#(2)$$

As for the time consumption, we record the rounded average time it takes for different models to complete multiple equal classification tasks. In real-world applications, sentiment classification models need to process large volumes of data within short timeframes. Thus, understanding the model's time overhead is crucial for selecting models suitable for specific application scenarios.

**Table 2.** Result of different models for original task.

| Method | Model | Accuracy | Weighted F1 Score | Weighted Recall | Time(s) |
|---|---|---|---|---|---|
| Machine Learning | Bayes+MLE | 0.023 | 0.01 | 0.03 | 7 |
| | Bayes+KNN | 0.009 | 0.01 | 0.02 | 79 |
| | SVM (linear) | 0.010 | 0.01 | 0.01 | 5 |
| | SVM (polynomial) | 0.016 | 0.03 | 0.02 | 6 |
| | SVM (rbf) | 0.015 | 0.01 | 0.02 | 6 |
| Natural Language Processing | BERT | 0.42 | 0.57 | 0.61 | 408 |
| | RoBERTa | 0.45 | 0.60 | 0.64 | 421 |
| | LLM(GPT) | 0.02 | 0.013 | 0.05 | >2000 |

**Table 3.** RoBERTa: results based on GoEmotions taxonomy.

| Emotion | Precision | Recall | F1 |
|---|---|---|---|
| admiration | 0.69 | 0.78 | 0.74 |
| amusement | 0.74 | 0.88 | 0.8 |
| anger | 0.53 | 0.5 | 0.51 |
| annoyance | 0.4 | 0.4 | 0.4 |
| approval | 0.43 | 0.38 | 0.4 |
| caring | 0.42 | 0.44 | 0.43 |
| confusion | 0.37 | 0.54 | 0.44 |
| curiosity | 0.48 | 0.65 | 0.55 |
| desire | 0.54 | 0.6 | 0.57 |
| disappointment | 0.41 | 0.31 | 0.35 |
| disapproval | 0.45 | 0.43 | 0.44 |
| disgust | 0.51 | 0.38 | 0.44 |
| embarrassment | 0.8 | 0.46 | 0.58 |
| excitement | 0.38 | 0.35 | 0.37 |
| fear | 0.75 | 0.56 | 0.64 |
| gratitude | 0.93 | 0.89 | 0.91 |
| grief | nan | 0.0 | nan |
| joy | 0.5 | 0.55 | 0.52 |
| love | 0.7 | 0.89 | 0.78 |
| nervousness | 0.67 | 0.29 | 0.4 |
| optimism | 0.64 | 0.57 | 0.61 |
| pride | 1.0 | 0.13 | 0.24 |
| realization | 0.4 | 0.23 | 0.29 |
| relief | nan | 0.0 | nan |
| remorse | 0.7 | 0.82 | 0.76 |
| sadness | 0.45 | 0.63 | 0.52 |
| surprise | 0.53 | 0.53 | 0.53 |
| neutral | 0.62 | 0.77 | 0.68 |
| **macro-average** | **0.58** | **0.5** | **0.53** |
| **std** | **0.17** | **0.24** | **0.17** |

**Table 4.** RoBERTa: results based on sentiment-grouped data.

| Sentiment | Precision | Recall | F1 |
|---|---|---|---|
| ambiguous | 0.57 | 0.60 | 0.59 |
| negative | 0.70 | 0.64 | 0.67 |
| neutral | 0.62 | 0.77 | 0.68 |
| positive | 0.82 | 0.83 | 0.82 |
| **macro-average** | **0.68** | **0.71** | **0.69** |
| **std** | **0.10** | **0.09** | **0.09** |

**Table 5.** RoBERTa: results based on Ekman's taxomony.

| Ekman Emotion | Precision | Recall | F1 |
|---|---|---|---|
| anger | 0.59 | 0.55 | 0.57 |
| disgust | 0.51 | 0.38 | 0.44 |
| fear | 0.76 | 0.54 | 0.63 |
| joy | 0.82 | 0.83 | 0.82 |
| neutral | 0.62 | 0.77 | 0.68 |
| sadness | 0.64 | 0.60 | 0.62 |
| surprise | 0.57 | 0.60 | 0.59 |
| **macro-average** | **0.65** | **0.61** | **0.62** |
| **std** | **0.10** | **0.14** | **0.11** |

**Analysis** The experimental findings illuminate significant disparities between machine learning models and the robust baseline established by Google's previous experiments (Table 2). Such deviations were anticipated, given the inherent limitations of these algorithms in adapting specifically to sentiment classification tasks. Despite these challenges, there are promising indications of the capacity of machine learning models to narrow the gap with the baseline, especially in scenarios characterized by a limited number of emotion labels. This suggests the potential for further refinement and optimization of machine learning techniques for sentiment analysis tasks. Moreover, the prolonged processing time observed in NLP models can be attributed to their inherent complexity, stemming from the intricate linguistic structures and semantic nuances they must contend with. Despite this computational overhead, NLP models demonstrate notable efficacy in capturing and interpreting subtle contextual cues, which are essential for accurate sentiment classification. Notably, the RoBERTa model emerges as the frontrunner among the evaluated models (Table 3), showcasing exceptional performance across a diverse range of classification tasks encompassing more than twenty emotion labels.

Compared to the baseline results obtained by the Google team for the original task (macro-average F1 score of 0.46), we observed a notable enhancement in the performance of the RoBERTa model in the 28-class classification task (macro-average F1 score of 0.53). Furthermore, across the original task, emotion-grouped task, or Ekman classification criteria, RoBERTa exhibited improvements in precision. However, there remains room for improvement in recall metrics.

The dataset exhibits varying levels of performance across different labels. Notably, certain labels, such as "gratitude" and "amusement," consistently achieve high F1 scores, often exceeding 0.8. However, the performance of certain labels is notably poorer. This discrepancy may be attributed, in part, to the limited number of training samples available for certain emotion categories. It should be noted that the occurrences of "nan" in Table 3 are due to insufficient data, resulting in instances where the sum of true positives and false positives for certain categories equals zero. For instance, the "relief" label is associated with significantly fewer training instances, which can contribute to its comparatively lower performance.

Let us delve into a more specific and in-depth analysis. RoBERTa's proficiency in sentiment analysis tasks is underpinned by its comprehensive pre-training methodology, architectural enhancements, and iterative refinement strategies. Firstly, RoBERTa's extensive pre-training regimen, characterized by larger-scale corpora and prolonged training durations, facilitates a deeper understanding of linguistic nuances and contextual intricacies. The utilization of dynamic masking strategies during pre-training fosters nuanced comprehension of contextual relationships within text, enhancing its efficacy in sentiment classification tasks.

Furthermore, RoBERTa's iterative refinement process and fine-tuning strategies significantly contribute to its performance in sentiment analysis. Through continuous training on both sentence-level and token-level pre-training objectives, RoBERTa effectively hones its language understanding capabilities, thereby facilitating more precise sentiment classification. This multi-faceted pre-training approach equips RoBERTa with a comprehensive grasp of semantic subtleties and syntactic structures, enabling it to discern subtle emotional nuances within text and make more accurate sentiment predictions.

Moreover, RoBERTa's architectural enhancements, such as increased model depth and wider context window, bolster its capacity to capture long-range dependencies and contextual nuances crucial for accurate sentiment analysis. By incorporating deeper transformer architectures and leveraging larger context windows during pre-training, RoBERTa effectively captures intricate linguistic patterns and contextual cues, enhancing its ability to discern nuanced emotional expressions and improve sentiment classification accuracy.

In contrast, the shortcomings of GPT in sentiment analysis tasks stem primarily from its unidirectional, generative nature and pre-training task selection. GPT's reliance on single-directional autoregressive language modeling may limit its ability to capture bidirectional contextual dependencies essential for accurate sentiment analysis. Furthermore, GPT's pre-training objective, focused on predicting the next word in a sequence, may not effectively capture the nuances of sentiment expression and contextual understanding required for sentiment analysis tasks. Additionally, GPT's architecture may be less adept at capturing long-range dependencies and contextual nuances compared to RoBERTa, further impeding its performance in sentiment analysis tasks.

In addressing the multifaceted considerations regarding our study, it is crucial to recognize the trade-off between performance and cost, particularly in the context of hardware limitations. Despite these challenges, we conducted meticulous experiments to derive reliable averaged results, ensuring the credibility of our research while navigating the constraints associated with hardware limitations. Moreover, the observed inefficiency and variability in the sentiment classification timing of GPT interfaces stem from various factors such as network latency, server load, and computational resource allocation. These factors converge to create instability and unpredictability in interface responsiveness, consequently impacting the stability of sentiment classification effectiveness. While our research did not extensively explore individual model computational complexities, our experimental design and result analysis provided a comprehensive assessment of BERT-like models, focusing on evaluating their classification accuracy and precision. Through comparisons of performance metrics among models like BERT and RoBERTa, we discerned a trade-off between computational overhead and model performance. Additionally, it is noteworthy that although traditional machine learning methods tested in our study exhibited minimal time overhead, their performance was significantly inferior, representing another form of trade-off in terms of performance versus computational cost. Despite the challenges posed by hardware limitations, the acceptable computational overhead of BERT-like models underscores their value in achieving robust performance in sentiment classification tasks.

## 4 Conclusion

In conclusion, this study innovatively expands the scope of model evaluation by encompassing a diverse range of machine learning and natural language processing models tailored to the GoEmotions dataset. Notably, our research has significantly broadened the horizons of model assessment, setting a new standard for evaluating sentiment classification tasks within Reddit platform comments. Through meticulous experimentation, RoBERTa emerged as the standout performer, attributed to its exceptional contextual comprehension, extensive pretraining, and fine-tuning capabilities. Looking ahead, future investigations will delve deeper into disentangling experiments aimed at unraveling the underlying principles of models like RoBERTa, thus enriching our understanding of their mechanisms. Moreover, the emergence of increasingly sophisticated fusion models holds the potential to revolutionize sentiment classification by harnessing the synergies inherent in multiple advanced models. These advancements are poised to propel the frontiers of sentiment analysis, facilitating deeper insights into user behavior and preferences across digital platforms. Thus, our study not only contributes to advancing the field of sentiment analysis but also lays the groundwork for future research endeavors aimed at unlocking the full potential of machine learning and natural language processing in understanding and interpreting human emotions in digital discourse.

## References

1. Carlo Strapparava and Rada Mihalcea. 2007. SemEval-2007 task 14: Affective text. In Proceedings of the Fourth International Workshop on Semantic Evaluations (SemEval-2007):70–74, Prague, Czech Republic. Association for Computational Linguistics.
2. Hassan Saif, Miriam Fernández, Yulan He and Harith Alani (2013). Evaluation datasets for Twitter sentiment analysis: a survey and a new dataset, the STS-Gold. In: 1st Interantional Workshop on Emotion and Sentiment in Social and Expressive Media: Approaches and Perspectives from AI (ESSEM 2013), 3 Dec 2013, Turin, Italy.
3. Paul Ekman. 1992. An argument for basic emotions. Cognition & Emotion, 6(3-4):169–200.
4. Ortony, A. (2022). Are All "Basic Emotions" Emotions? A Problem for the (Basic) Emotions Construct. Perspectives on Psychological Science, 17(1):41-61.

5. Dorottya Demszky, Dana Movshovitz-Attias, Jeongwoo Ko, Alan S. Cowen, Gaurav Nemade, Sujith Ravi (2020) GoEmotions: A Dataset of Fine-Grained Emotions. Proceedings of the 58th Annual Meeting of the Association for Computational Linguistics:4040–4054.
6. Thomas Wolf, Lysandre Debut, Victor Sanh, Julien Chaumond, Clement Delangue, Anthony Moi, Pierric Cistac, Tim Rault, Rémi Louf, Morgan Funtowicz, Joe Davison, Sam Shleifer, Patrick von Platen, Clara Ma, Yacine Jernite, Julien Plu, Canwen Xu, Teven Le Scao, Sylvain Gugger, Mariama Drame, Quentin Lhoest, Alexander M. Rush (2019) HuggingFace's Transformers: State-of-the-art Natural Language Processing. Proceedings of the 2020 Conference on Empirical Methods in Natural Language Processing: System Demonstrations:38–45.
7. Jacob Devlin, Ming-Wei Chang, Kenton Lee, and Kristina Toutanova (2019) Bert: Pre-training of deep bidirectional transformers for language understanding. Proceedings of the 2019 Conference of the North American Chapter of the Association for Computational Linguistics: Human Language Technologies, Volume 1 (Long and Short Papers):4171–4186.
8. Yinhan Liu, Myle Ott, Naman Goyal, Jingfei Du, Mandar Joshi, Danqi Chen, Omer Levy, Mike Lewis, Luke Zettlemoyer, Veselin Stoyanov (2019) RoBERTa: A Robustly Optimized BERT Pretraining Approach. arXiv:1907.11692.